\documentclass[journal,twoside,web]{ieeecolor}
\usepackage{generic}
\usepackage{cite}

\makeatletter
\let\NAT@parse\undefined
\makeatother

\usepackage[
    colorlinks=true,
    linkcolor=blue,
    citecolor=blue,
    urlcolor=blue
]{hyperref}

\usepackage{makecell}
\usepackage{array}
\usepackage{amsmath,amssymb,amsfonts}
\usepackage{adjustbox}
\usepackage{booktabs}
\usepackage{graphicx}
\usepackage{algorithm,algorithmic}
\usepackage{makecell}
\usepackage{placeins}
\usepackage{textcomp}
\usepackage{tabularx}
\usepackage{graphicx}
\usepackage{multirow}
\usepackage{booktabs}
\newcommand{\doi}[1]{\href{https://doi.org/#1}{#1}}

\def\BibTeX{{\rm B\kern-.05em{\sc i\kern-.025em b}\kern-.08em
    T\kern-.1667em\lower.7ex\hbox{E}\kern-.125emX}}
\begin{document}
\title{TransECG-SSL: Hybrid Self-Supervised Representation Learning for ECG Arrhythmia Classification}
\author{
Ahtisham Ayyub, \IEEEmembership{Student Member, IEEE},
Fasih Haider, \IEEEmembership{Senior Member, IEEE},
Muhammad Rehan Usman, \IEEEmembership{Member, IEEE},
Saturnino Luz, \IEEEmembership{Member, IEEE},
Christos Politis, \IEEEmembership{Senior Member, IEEE},
and Muhammad Arslan Usman, \IEEEmembership{Member, IEEE}
\thanks{Manuscript submitted for review on August 28, 2026.
This work has been submitted to the IEEE for possible publication.
Copyright may be transferred without notice, after which this version may no longer be accessible.}
\thanks{Ahtisham Ayyub, Muhammad Rehan Usman, Christos Politis, and
Muhammad Arslan Usman are with the School of Computer Science and
Mathematics, Kingston University, London, United Kingdom
(e-mail: \{a.ayyub, r.usman, c.politis, m.usman\}@kingston.ac.uk).}
\thanks{Fasih Haider and Saturnino Luz are with the Usher Institute,
University of Edinburgh, Edinburgh, United Kingdom
(e-mail: \{fasih.haider, s.luz\}@ed.ac.uk). FH and SL would like to acknowledge UKRI under grant number 10102226.}
}

\maketitle

\begin{abstract}
Cross-dataset generalisation remains a major challenge for automated electrocardiogram (ECG) arrhythmia classification because of differences in patient populations, acquisition protocols, annotation practices, and class distributions. Learning representations that remain informative across independent datasets could reduce reliance on dataset-specific labelled data. We propose TransECG-SSL, a self-supervised framework for transferable heartbeat representation learning from unlabelled ECG data. A Transformer encoder is pretrained using a hybrid objective combining masked signal reconstruction and contrastive learning, then frozen and evaluated by linear probing under patient-disjoint evaluation. Experiments across three public ECG databases and seven source--target settings show that Hybrid SSL consistently outperforms randomly initialised Transformers and achieves higher balanced accuracy than architecture-matched source-supervised pretraining. Reconstruction-only SSL remains competitive with Hybrid SSL, whereas contrastive-only pretraining performs substantially worse, highlighting the importance of masked reconstruction for transferability. Cross-dataset transfer nevertheless reduces balanced accuracy relative to target-specific self-supervised pretraining, while minority arrhythmia classes remain challenging under severe class imbalance. These findings demonstrate the potential of label-efficient self-supervised learning for cross-dataset ECG representation transfer while highlighting persistent limitations from domain shift and class imbalance.
\end{abstract}

\begin{IEEEkeywords}
Arrhythmia classification, Cross-dataset generalisation, Electrocardiogram (ECG), Self-supervised learning, Transfer learning, Transformer, Representation learning, Domain shift 
\end{IEEEkeywords}

\section{Introduction}
\label{sec:introduction}
\IEEEPARstart{E}{lectrocardiography} (ECG) is widely used to assess cardiac rhythm, and the increasing volume of ECG data generated in clinical, ambulatory and wearable-monitoring settings has motivated substantial research on automated arrhythmia classification. Although Deep-learning (DL) methods can learn discriminative representations directly from ECG signals; reported performance remains difficult to compare because studies vary substantially in datasets, preprocessing, architectures, evaluation protocols, and metrics \cite{hong2020opportunities,ayyub2025comprehensive}. Reviews also report a high dependency on MIT-BIH, a greater use of intra-patient rather than inter-patient evaluation, and less favourable class-sensitive metrics than headline accuracy \cite{xiao2023deep}. These concerns reinforce the need for patient-independent evaluation and careful interpretation of high reported performance \cite{silva2026systematic}.

Heartbeat-level evaluation is significantly affected by patient overlap because individual patients contribute many correlated beats; patient-disjoint partitioning therefore provides a stricter test of generalisation. Severe class imbalance adds a second difficulty, since normal beats dominate many arrhythmia datasets and minority classes may be poorly recognised even when overall accuracy is high \cite{han2022hctnet}. 

Another challenge is the dependence of supervised ECG models on time-consuming and expensive expert-labelled data. Self-supervised learning (SSL) offers a way to exploit unlabelled ECG signals for representation learning. ECG SSL has been investigated through contrastive approaches \cite{chen2021clecg}, broader self-supervised representation learning \cite{mehari2022self}, masked reconstruction \cite{zhang2022maefe}, out-of-distribution evaluation \cite{soltanieh2023distribution}, and larger hybrid or foundation-model frameworks \cite{mckeen2025ecg}. These strands motivate a controlled comparison of reconstructive and contrastive objectives for transferable heartbeat representations.

Transfer learning can reduce dependence on target-domain labels \cite{kachuee2018ecg,weimann2021transfer}, but transfer across independent ECG databases is complicated by differences in patient populations, acquisition conditions, annotation practices, and class distributions; reported performance can also depend materially on partitioning choices \cite{cao2023ecg}. Recent approaches address such shifts through domain adaptation \cite{imtiaz2024cross}, domain generalisation \cite{ballas2022domain}, and domain-invariant or multi-dataset learning \cite{li2025research,kim2026enhancing}. However, these approaches differ in the information available during training, for instance transfer learning typically uses labelled target samples for fine-tuning, domain adaptation exploits unlabelled target data to align source and target representations, whereas domain generalisation and domain-invariant learning aim to learn representations that remain robust to unseen target distributions without target-specific adaptation.
These observations motivate a narrower representation-transfer question: whether an ECG encoder pretrained on one arrhythmia database can remain fixed and provide useful heartbeat representations on an independent target database when only a lightweight target classifier is fitted. Such an evaluation differs from target-domain adaptation because the source-pretrained encoder is not updated using target data, and it differs from conventional transfer learning in which the pretrained encoder is fine-tuned on the target task. It also requires architecture-matched comparisons between self-supervised objectives so that any apparent advantage is not attributable simply to differences in encoder capacity or downstream evaluation.
To investigate this question, this study proposes TransECG-SSL, a hybrid self-supervised framework for ECG heartbeat representation learning.

The main contributions of this study are: 
\textit{(i)} TransECG-SSL, a Transformer-based framework combining masked reconstruction and contrastive learning to learn transferable heartbeat representations from unlabelled ECG data; 
\textit{(ii)} systematic evaluation of frozen representation transfer across seven within- and cross-dataset settings spanning MIT-BIH, INCART, and SVDB under patient-disjoint evaluation; 
\textit{(iii)} architecture-matched comparison with reconstruction-only, contrastive-only, source-supervised, and random-encoder baselines under a common linear-probing protocol; and 
\textit{(iv)} analysis of cross-dataset generalisation and class-specific performance, demonstrating transferable self-supervised representations while identifying domain shift and minority-class recognition as persistent limitations.

\section{Related Work}
\label{sec:related}

\subsection{Patient-Independent Arrhythmia Classification and Class Imbalance}

Patient-independent evaluation has been adopted in several heartbeat classification studies to prevent beats from same individual contributing to both training and testing partitions. Existing inter-patient approaches include single-lead classification, dual-attention networks, residual convolutional architectures, weighted classifiers, symbolic representations, and morphological methods \cite{dias2021arrhythmia,lyu2023automated,li2022inter,xu2023inter,niu2019inter,zakaria2024morphological}. Reviews nevertheless indicate that inter-patient evaluation remains less common and generally more demanding than intra-patient evaluation \cite{xiao2023deep,silva2026systematic}.

Patient variability is closely coupled with severe class imbalance. In commonly used heartbeat datasets, normal beats substantially outnumber several abnormal classes, creating conditions in which optimisation can favour majority-class recognition. HCTNet, for example, explicitly addresses the combination of inter-patient variability and imbalanced arrhythmia data \cite{han2022hctnet}. Other studies have approached imbalance through adversarial data synthesis, augmentation, adaptive training, class-aware weighting, or specialised rare-beat learning \cite{rafi2022heartnet,tiffany2026robust,dehghani2026rare}. Yuan et al.\ further considered imbalance jointly with domain adaptation, using target-oriented augmentation and class-aware mechanisms to address changes in both class distribution and source--target domain characteristics \cite{yuan2023target}. These studies demonstrate that minority-class performance cannot be inferred reliably from overall accuracy alone and motivate class-sensitive metrics and classwise analysis.

Most of this literature, however, focuses on improving the final supervised classifier within a particular dataset or evaluation protocol. Attention mechanisms, class weighting, synthetic data generation, and adaptive sampling may improve minority-class decision boundaries, but they do not directly determine whether the underlying representation itself remains transferable across independent databases. The present study therefore treats patient-disjoint partitioning and class imbalance primarily as constraints on representation-transfer evaluation rather than introducing a specialised imbalance-correction algorithm.

\subsection{Self-Supervised ECG Representation Learning}

ECG SSL spans contrastive, reconstructive, and hybrid approaches. Contrastive methods include CLECG \cite{chen2021clecg} and broader ECG-specific self-supervised pretraining \cite{mehari2022self}, with recent extensions exploring multimodal objectives, physiology-aware augmentations, temporal--spatial learning, and multi-lead classification \cite{hu2025novel,maghsoodi2026domain,chen2025temporal,ma2026self}. RhythmX additionally combines contrastive representation learning with heartbeat classification and post-hoc interpretability \cite{abdullah2026rhythmx}.

Masked reconstruction provides a complementary learning signal. MaeFE demonstrated ECG-specific masked-autoencoder pretraining through transfer, linear probing, and fine-tuning \cite{zhang2022maefe}, while subsequent work has combined masked representations with clustering or discriminative objectives \cite{zhang2025sequential} and Transformer-based downstream evaluation \cite{chatterjee2026toward}. These studies indicate that reconstruction alone can provide a strong ECG representation-learning signal.

Hybrid approaches, on the other hand, combine reconstruction with contrastive alignment, physiological priors, multimodal learning, or large-scale pretraining \cite{maghsoodi2026domain,zhang2025sequential,mckeen2025ecg}; ECG-FM, for example, combines masked reconstruction and contrastive learning \cite{mckeen2025ecg}. Because contrastive performance also depends on view construction and augmentation \cite{chen2020simple,maghsoodi2026domain}, controlled objective-level comparisons are important.
Hybrid SSL, reconstruction-only SSL, and contrastive-only SSL use the same unlabelled source training partitions and the same Transformer encoder architecture, allowing differences in frozen downstream representations to be attributed primarily to the pretraining objective. This distinction is important because prior ECG SSL studies differ substantially in dataset, encoder architecture, augmentation policy, downstream classifier, and fine-tuning strategy, making numerical comparisons across publications insufficient for determining whether reconstruction, contrastive learning, or their combination is responsible for improved transfer.

\subsection{ECG Transfer Learning and Cross-Dataset Generalisation}

ECG transfer-learning studies have shown that pretrained representations can be reused across cardiac tasks and can reduce target-label requirements \cite{kachuee2018ecg,weimann2021transfer,mavaddati2025ecg,cao2023ecg}, although conventional transfer learning generally permits target fine-tuning.

Methods addressing ECG domain shift include unsupervised domain adaptation \cite{imtiaz2024cross,yuan2023target}, domain-adaptive normalisation \cite{niloy2025lead}, domain generalisation \cite{ballas2022domain}, domain-invariant representation learning \cite{li2025research}, unified multi-dataset training \cite{kim2026enhancing}, and distributionally adaptive frameworks \cite{wu2025one}. These approaches may use target-domain information, multiple training domains, or explicit adaptation mechanisms. Frozen representation transfer asks a narrower question: whether a source-pretrained encoder can remain unchanged while supporting a lightweight classifier on an independent target dataset.

Existing literature therefore provides substantial evidence for patient-independent ECG evaluation, self-supervised representation learning, transfer learning, and explicit methods for addressing domain shift. However, these strands have generally been investigated using heterogeneous datasets, label spaces, architectures, optimisation procedures, and assumptions about access to target-domain data. Cross-publication performance comparisons consequently cannot isolate whether differences arise from the representation-learning objective, model architecture, patient partitioning, target adaptation, or downstream classifier.

Accordingly, a controlled evaluation is needed in which encoder architecture and downstream probing are held fixed while alternative representation-learning objectives are compared under patient-disjoint transfer. This study addresses that gap through TransECG-SSL and architecture-matched baselines.

\section{Methodology}
\label{sec:method}

\subsection{Overview of TransECG-SSL}
\label{subsec:overview}

\begin{figure}
    \centering
    \includegraphics[width=\linewidth]{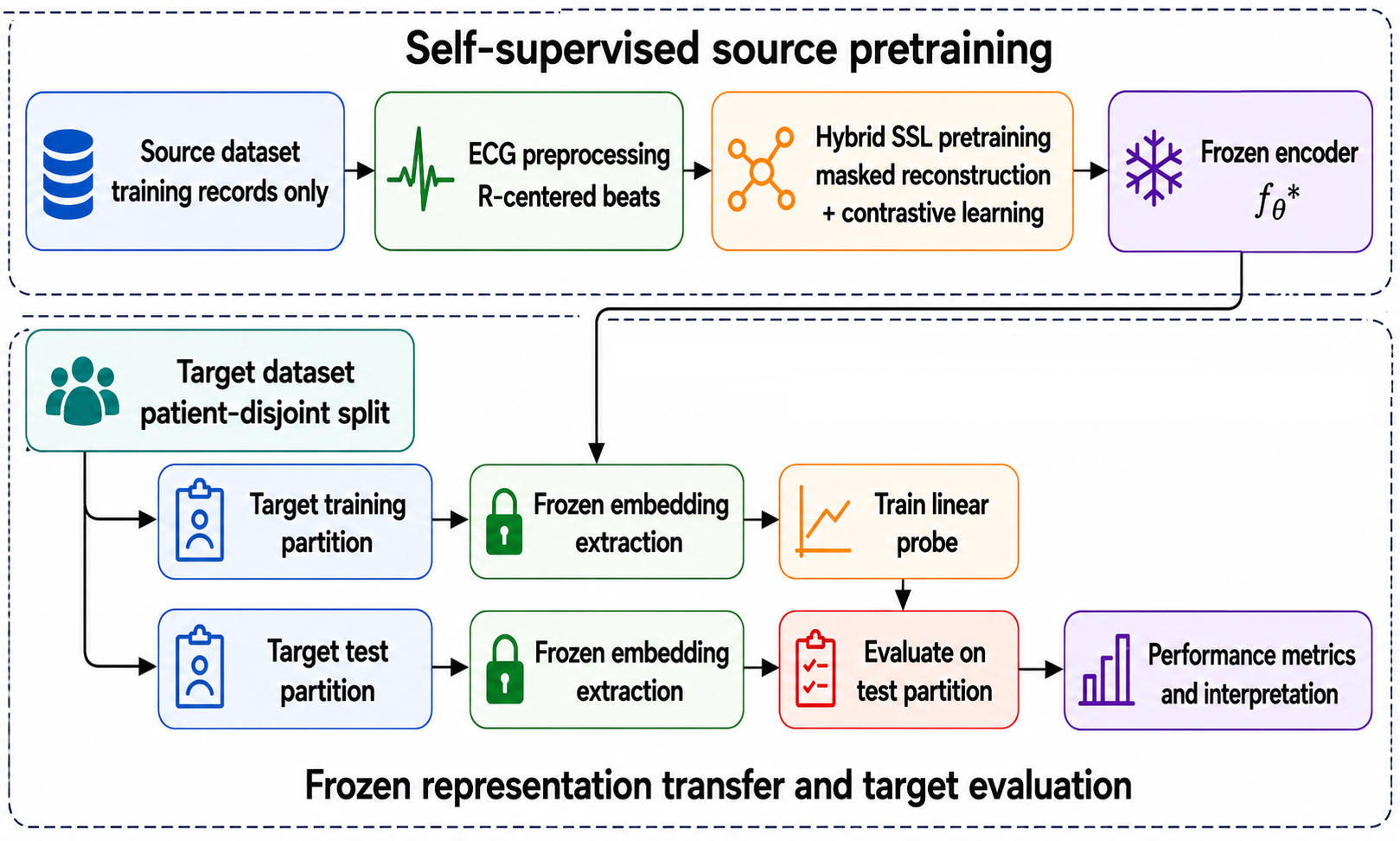}
    \caption{Proposed Methodology}
    \label{fig:system}
\end{figure}

Figure~\ref{fig:system} summarises the proposed TransECG-SSL framework. The method is designed to evaluate whether self-supervised ECG representations learned from one arrhythmia database can transfer to independent target databases under patient-disjoint evaluation. The framework follows three main stages. First, a Transformer-based ECG encoder is pretrained using only the training records of a source dataset. The self-supervised objective combines masked ECG reconstruction with contrastive representation learning, allowing the encoder to learn morphology-preserving and augmentation-invariant heartbeat representations without using class labels during pretraining. Second, the pretrained encoder is frozen and used to extract clean heartbeat embeddings separately from the target training and test partitions. Third, a linear probe is trained on embeddings from the target training partition and evaluated on embeddings from the patient-disjoint target test partition.

This design separates representation learning from supervised classification and avoids direct optimisation of the encoder on the target test set. Within-dataset experiments provide target-matched reference performance, whereas cross-dataset experiments evaluate the transferability of frozen representations across independent ECG databases.

\subsection{ECG Datasets and Preprocessing }

Three publicly available ECG arrhythmia databases were used to evaluate TransECG-SSL: the MIT-BIH Arrhythmia Database \cite{moody2001impact}, the St Petersburg INCART 12-lead Arrhythmia Database \cite{incart}, and the MIT-BIH Supraventricular Arrhythmia Database (SVDB) \cite{greenwald1990improved}. We adopted the five ANSI/AAMI heartbeat superclasses: normal beats (N), supraventricular ectopic beats (S), ventricular ectopic beats (V), fusion beats (F), and unknown or paced beats (Q) \cite{AAMI_EC57_2012}. This grouping is also consistent with the widely used heartbeat-classification protocol of de Chazal et al. \cite{de2004automatic}. Moreover, SVDB was evaluated using the shared N/S/V classes only, because F and Q beats were severely under-represented after annotation mapping and beat segmentation. Unknown or non-beat annotation symbols were excluded rather than being forced into the Q class. Table~\ref{tab:final_experiment_matrix} summarises the class setup.

All datasets were processed using the same beat-centred preprocessing pipeline to reduce dataset-specific preprocessing bias before self-supervised learning and transfer evaluation. Table~\ref{tab:preprocessing_params} summarises the preprocessing and beat-segmentation settings used throughout all experiments.

\begin{table}[t]
\centering
\caption{ECG preprocessing and beat-segmentation parameters used in TransECG-SSL.}
\label{tab:preprocessing_params}
\small
\renewcommand{\arraystretch}{1.15}
\begin{tabularx}{\linewidth}{@{}p{2.8cm}X@{}}
\toprule
\textbf{Component} & \textbf{Setting} \\
\midrule
Input source & WFDB ECG records and reference beat annotations. \\
ECG lead & First available ECG lead was used consistently across datasets. \\
Bandpass filtering & Fourth-order Butterworth bandpass filter from 0.5--40 Hz using zero-phase second-order-section filtering. \\
Resampling & Continuous ECG signals were resampled to 256 Hz before beat extraction. R-peak locations were mapped to the resampled signal index space. \\
Beat alignment & Beats were centred on the annotated R-peak. \\
Window length & 256 samples per beat. \\
Window position & 128 samples before and 128 samples after the R-peak. \\
Boundary handling & Beats were discarded if the complete 256-sample window could not be extracted. No padding was used. \\
Normalisation & Per-beat z-score normalisation. \\
Output representation & Fixed-length single-channel ECG beat segments used for SSL pretraining, embedding extraction, and linear probing. \\
\bottomrule
\end{tabularx}
\end{table}

\subsection{Patient-Disjoint Class-Aware Splitting}

All experiments used fixed patient-disjoint train/test partitions. Recordings from the same patient were assigned jointly to one partition, preventing patient overlap between training and testing. For SVDB, each recording corresponds to a distinct patient; therefore, patient- and record-level partitioning are equivalent.

To improve rare-class coverage, patient-level AAMI class counts were used to construct class-aware splits with an approximately 2:1 train/test patient ratio. Split generation used only patient identifiers and class counts and was completed before model training or evaluation. The resulting partitions, which were fixed across all five seeded runs, are summarised in Table~\ref{tab:classaware_splits}. INCART retained the extremely sparse Q class to preserve the five-class AAMI setting; its effect on the aggregate metrics was therefore examined separately in the sensitivity analysis reported in Table~\ref{tab:incart_q_sensitivity}.

\begin{table}[t]
\centering
\caption{Patient-disjoint class-aware train/test splits used in the final experiments.}
\label{tab:classaware_splits}

\small
\resizebox{\linewidth}{!}{%
\begin{tabular}{llrrrrrrrr}
\toprule
\textbf{Dataset} & \textbf{Split} & \textbf{Pts.} & \textbf{Recs.} &
\textbf{N} & \textbf{S} & \textbf{V} & \textbf{F} & \textbf{Q} & \textbf{Total} \\
\midrule

MIT & Train & 31 & 31 & 56,914 & 1,861 & 4,802 & 435 & 6,038 & 70,050 \\
    & Test  & 16 & 17 & 33,668 & 920 & 2,433 & 367 & 2,132 & 39,520 \\

\midrule

INC & Train & 21 & 51 & 104,015 & 1,333 & 14,351 & 147 & 4 & 119,850 \\
    & Test  & 11 & 24 & 49,581 & 625 & 5,655 & 72 & 2 & 55,935 \\

\midrule

SVDB & Train & 52 & 52 & 108,676 & 8,167 & 6,656 & -- & -- & 123,499 \\
     & Test  & 26 & 26 & 53,575 & 4,028 & 3,282 & -- & -- & 60,885 \\

\bottomrule
\end{tabular}%
}

\vspace{0.3em}
\parbox{\linewidth}{%
\scriptsize
\textit{Note:} MIT: MIT-BIH; INC: INCART; Pts.: patients; Recs.: ECG records;
Total: total beats. SVDB evaluation was restricted to the N/S/V classes;
therefore, F and Q are not reported. Owing to class imbalance, an N-only
majority classifier would achieve accuracies of .852, .886, and .880 on the
MIT-BIH, INCART, and SVDB test sets, respectively, while its balanced accuracy
would remain at chance level (.200 for the five-class tasks and .333 for SVDB).
}

\end{table}

\subsection{Self-Supervised Encoder and Hybrid Objective}

TransECG-SSL uses a one-dimensional Transformer encoder to learn fixed-length ECG heartbeat representations. Each input beat $x_i \in \mathbb{R}^{T}$, where $T=256$, is first projected into a token sequence using a convolutional stem. A learnable class token is then prepended to the sequence and sinusoidal positional encoding is added. The sequence is passed through a Transformer encoder, based on the self-attention architecture \cite{vaswani2017attention}. The class-token output is projected to a compact embedding vector:
\begin{equation}
    z_i = f_{\theta}(x_i), \qquad z_i \in \mathbb{R}^{p}
\end{equation}
where $f_{\theta}$ is the ECG encoder and $p=128$ is the dimension of the embedding.

The encoder was pretrained using a hybrid self-supervised objective that combines masked ECG reconstruction and contrastive representation learning. For masked reconstruction, a binary temporal mask $m_i \in \{0,1\}^{T}$ was sampled using contiguous time blocks. The masked input was defined as:
\begin{equation}
    \tilde{x}_i = (1-m_i)\odot x_i + m_i \odot v
\end{equation}
where $v$ is the mask replacement value and $\odot$ denotes element-wise multiplication. The masked beat was passed through the encoder and decoder:
\begin{equation}
    z_i^{m}=f_{\theta}(\tilde{x}_i),
    \qquad
    \hat{x}_i=g_{\phi}(z_i^{m})
\end{equation}
where $g_{\phi}$ denotes the reconstruction decoder. The reconstruction loss was computed only over masked positions:
\begin{equation}
    \mathcal{L}_{\mathrm{rec}}
    =
    \frac{1}{\sum_i \sum_t m_{i,t}}
    \sum_i \sum_{t=1}^{T}
    m_{i,t}
    \left(\hat{x}_{i,t}-x_{i,t}\right)^2
\end{equation}

For the contrastive branch, a weakly augmented view $x_i^{+}$ of the same beat was generated using additive Gaussian noise and amplitude scaling. The augmented view was passed through the same encoder:
\begin{equation}
    z_i^{+}=f_{\theta}(x_i^{+})
\end{equation}
The masked-view and augmented-view embeddings were then mapped to a contrastive projection space:

\begin{equation}
    h_i^{m}=q_{\psi}(z_i^{m}),
    \qquad
    h_i^{+}=q_{\psi}(z_i^{+})
\end{equation}
where $q_{\psi}$ is the projection head. The contrastive loss used cosine similarity:

\begin{equation}
    \operatorname{sim}(a,b)
    =
    \frac{a^{\top}b}{\|a\|_2\|b\|_2}
\end{equation}

For a batch of $B$ beats, the contrastive objective was computed using the normalized temperature-scaled cross-entropy loss (NT-Xent) from the SimCLR formulation \cite{chen2020simple}:

\begin{equation}
    \mathcal{L}_{\mathrm{con}}
    =
    -\frac{1}{2B}
    \sum_{i=1}^{2B}
    \log
    \frac{
    \exp\left(\operatorname{sim}(h_i,h_{p(i)})/\tau\right)
    }{
    \sum_{\substack{j=1 \ j\neq i}}^{2B}
    \exp\left(\operatorname{sim}(h_i,h_j)/\tau\right)
    }
\end{equation}
where $p(i)$ indexes the positive counterpart of sample $i$, and $\tau$ is the temperature parameter.

The final hybrid SSL objective was:

\begin{equation}
    \mathcal{L}_{\mathrm{SSL}}
    =
    \lambda_{\mathrm{rec}}\mathcal{L}_{\mathrm{rec}}
    +
    \lambda_{\mathrm{con}}\mathcal{L}_{\mathrm{con}}
\end{equation}
In the final configuration, $\lambda_{\mathrm{rec}}=1.0$, $\lambda_{\mathrm{con}}=0.10$, and $\tau=0.20$ were used. Pretraining was performed on source-dataset training records only.

\subsection{Architecture-Matched Baselines and Supervised References}

To separate the effect of the training objective from encoder architecture, additional baselines were implemented using the same Transformer encoder as TransECG-SSL. The shared encoder contains a convolutional stem, class token, four Transformer layers with four attention heads, model dimension 256, feed-forward dimension 1024, dropout 0.1, and a 128-dimensional output embedding. This encoder contains 3,358,592 trainable parameters in all architecture-matched conditions.

The \emph{reconstruction-only SSL} condition uses the same masking configuration, reconstruction decoder, encoder architecture, source records, and optimisation schedule as Hybrid SSL, but removes the contrastive contribution by setting $\lambda_{\mathrm{con}}=0$. It therefore isolates the contribution of adding contrastive representation learning to masked reconstruction.

The \emph{contrastive-only SSL} condition uses the same Transformer encoder and 64-dimensional contrastive projection head, but removes masking and reconstruction entirely. Two independently generated weakly augmented views of each clean heartbeat are used as the positive pair. Both views use the same Gaussian-noise and amplitude-scaling augmentation strengths as Hybrid SSL, and the encoder is optimised solely with the NT-Xent loss using temperature $\tau=0.20$. As in Hybrid SSL, contrastive negatives are formed within each distributed-process batch rather than gathered across processes.

The \emph{source-supervised Transformer} uses the same encoder but replaces self-supervised pretraining with a temporary linear classification head trained using labelled source training records and ordinary multiclass cross-entropy. After source training, the classification head is discarded, the encoder is frozen, and target train/test embeddings are evaluated using exactly the same target linear-probe protocol as the SSL conditions. This provides an architecture-matched comparison between supervised and self-supervised source pretraining.

The \emph{random frozen Transformer} uses the identical encoder architecture with random initialisation and no encoder pretraining. It is frozen before target embedding extraction and serves as an architecture-matched lower-reference condition. Because no source-domain training occurs, this condition is target-specific rather than source-specific.

Finally, a \emph{target-supervised end-to-end Transformer} was trained from random initialisation using labelled target training records. The same Transformer encoder and a linear classification head were optimised jointly and evaluated directly on the corresponding disjoint target test records. This condition provides a target-supervised classification reference rather than a source-to-target representation-transfer ablation.

Hybrid SSL, reconstruction-only SSL, contrastive-only SSL, the source-supervised Transformer, and the target-supervised Transformer were trained for 50 epochs using AdamW, a learning rate of $10^{-3}$, weight decay of $10^{-4}$, and global batch size 256. All frozen-representation conditions were evaluated using the same unweighted target linear probe trained for 30 epochs with Adam, learning rate $10^{-3}$, and batch size 512. In these comparisons, no inverse-frequency class weighting or weighted sampling was used.

\subsection{Frozen Representation Transfer with Linear Probing}

After self-supervised pretraining, the reconstruction decoder and contrastive projection head were discarded. The pretrained encoder was frozen and used only as a feature extractor. For each target dataset, clean embeddings were extracted separately from the target training and test records:

\begin{equation}
    z_i = f_{\theta^{*}}(x_i)
\end{equation}
where $f_{\theta^{*}}$ denotes the pretrained frozen encoder and $z_i \in \mathbb{R}^{128}$ is the resulting heartbeat embedding. No masking or reconstruction operation was applied during embedding extraction.

A linear probe was then trained on the target training embeddings and evaluated on the disjoint target test embeddings:
\begin{equation}
    \hat{y}_i = \operatorname{softmax}(Wz_i+b)
\end{equation}
where $W$ and $b$ are the trainable parameters of the linear classifier. The encoder parameters $\theta^{*}$ remained fixed during probe training. The probe was optimised using ordinary cross-entropy loss:
\begin{equation}
    \mathcal{L}_{\mathrm{probe}}
    =
    -\frac{1}{N}
    \sum_{i=1}^{N}
    \sum_{c=1}^{C}
    \mathbf{1}(y_i=c)\log(\hat{y}_{i,c})
\end{equation}
In the main analysis, no class-weighted loss or weighted sampler was used because the objective was to evaluate the transferability of the frozen self-supervised representation rather than to optimise minority-class decision thresholds.

Table~\ref{tab:final_experiment_matrix} summarises the final frozen-transfer evaluation matrix. Three within-target experiments were used as target-matched references, while four cross-dataset experiments assessed representation transfer across independent ECG databases. Evaluated classes for each experiment setting have been placed in 'Evaluated classes' column.

\begin{table*}[t]
\centering
\caption{Frozen representation transfer and linear-probe evaluation matrix used in the final TransECG-SSL experiments.}
\label{tab:final_experiment_matrix}
\scriptsize
\renewcommand{\arraystretch}{1.15}
\begin{tabularx}{\textwidth}{@{}p{2.3cm}p{2.8cm}p{5.4cm}p{1.3cm}X@{}}
\toprule
\textbf{Experiment} & \textbf{SSL pretraining source} & \textbf{Frozen embedding extraction} & \textbf{Evaluated classes} & \textbf{Purpose} \\
\midrule
MIT-BIH $\rightarrow$ MIT-BIH & MIT-BIH training partition & MIT-BIH patient-disjoint training and test partitions & N/S/V/F/Q & Target-matched reference for MIT-BIH. \\
INCART $\rightarrow$ INCART & INCART training partition & INCART patient-disjoint training and test partitions & N/S/V/F/Q & Target-matched reference for INCART. \\
SVDB $\rightarrow$ SVDB & SVDB training partition & SVDB patient-disjoint training and test partitions & N/S/V & Target-matched reference for SVDB. \\
\midrule
MIT-BIH $\rightarrow$ INCART & MIT-BIH training partition & INCART patient-disjoint training and test partitions & N/S/V/F/Q & Cross-dataset transfer from MIT-BIH to INCART. \\
INCART $\rightarrow$ MIT-BIH & INCART training partition & MIT-BIH patient-disjoint training and test partitions & N/S/V/F/Q & Cross-dataset transfer from INCART to MIT-BIH. \\
MIT-BIH $\rightarrow$ SVDB & MIT-BIH training partition & SVDB patient-disjoint training and test partitions & N/S/V & Cross-dataset transfer from MIT-BIH to SVDB. \\
INCART $\rightarrow$ SVDB & INCART training partition & SVDB patient-disjoint training and test partitions & N/S/V & Cross-dataset transfer from INCART to SVDB. \\
\bottomrule
\end{tabularx}
\end{table*}

\subsection{Evaluation Metrics and Transfer Analysis}

Performance was evaluated on the held-out target test records for each source--target setting. Overall performance was assessed using accuracy, balanced accuracy (BA), macro-F1, macro-AUROC, macro-AUPRC, Matthews correlation coefficient (MCC) \cite{chicco2020advantages}, and Cohen's kappa \cite{kvaalseth1989note}. Given the substantial class imbalance in the evaluated datasets, BA and macro-F1 were used as the principal class-balanced metrics. BA represents the macro-average of class-wise recall, whereas macro-F1 gives equal weight to the F1 score of each class. Macro-AUROC and macro-AUPRC were used to assess one-vs-rest ranking performance across evaluable classes.

Class-specific performance was further examined using sensitivity, positive predictive value (PPV), F1 score, AUROC, and AUPRC. Row-normalised confusion matrices were used to identify systematic misclassification patterns, particularly for minority arrhythmia classes. All final experiments were repeated using five random seeds, and results are reported as mean $\pm$ sample standard deviation (SD).

To quantify cross-dataset transfer relative to each method's own within-target performance, target-normalised transfer gaps were calculated for Hybrid SSL, reconstruction-only SSL, and the source-supervised Transformer. For method $m$, source dataset $s \neq t$, target dataset $t$, metric $M$, and matched seed $k$, the target-normalised transfer gap was defined as:

\begin{equation}
G_{m,s\rightarrow t}^{(k)}
=
M_{m,s\rightarrow t}^{(k)}
-
M_{m,t\rightarrow t}^{(k)}.
\end{equation}

Thus, $G_{m,s\rightarrow t}^{(k)}<0$ indicates that a method performed worse after cross-dataset transfer than when the same method was trained within the target dataset, whereas a positive value indicates higher cross-dataset performance. MIT-BIH $\rightarrow$ INCART was referenced to INCART $\rightarrow$ INCART, INCART $\rightarrow$ MIT-BIH to MIT-BIH $\rightarrow$ MIT-BIH, and both SVDB-target transfer routes to SVDB $\rightarrow$ SVDB. The reference was always taken from the same representation-learning method as the corresponding cross-dataset result. 

Transfer gaps were calculated separately for balanced accuracy and macro-F1, the two principal class-balanced metrics, for each of the five matched seeds and then averaged across seeds. The resulting mean gaps were reported alongside the corresponding mean cross-dataset performance to distinguish absolute transfer performance from degradation relative to each method's own within-target reference.
\section{Results}
\label{sec:res}
\subsection{Overall Transfer Performance of Hybrid SSL}

Table~\ref{tab:hybrid_overall_transfer} summarises the overall performance of the proposed Hybrid SSL encoder across the seven source--target evaluation routes, comprising three within-target references and four cross-dataset transfers. In all experiments, the encoder was pretrained using unlabelled data from the source training partition only and then frozen before target-dataset embedding extraction. A linear probe was trained on embeddings from the target training partition and evaluated on embeddings from the patient-disjoint target test partition. Therefore, the reported results assess the transferability of the source-pretrained representation rather than end-to-end supervised optimisation on the target test set.

\begin{table*}[t]
\centering
\caption{Overall transfer performance of the proposed Hybrid SSL encoder.}
\label{tab:hybrid_overall_transfer}

\scriptsize
\renewcommand{\arraystretch}{1.05}
\setlength{\tabcolsep}{2.5pt}

\resizebox{0.99\textwidth}{!}{%
\begin{tabular}{lcccccccc}
\toprule
\textbf{Experiment} &
\textbf{Acc.} &
\textbf{BA} &
\textbf{A--BA gap} &
\textbf{F1} &
\textbf{AUROC} &
\textbf{AUPRC} &
\textbf{MCC} &
\textbf{$\kappa$} \\
\midrule

MIT $\rightarrow$ MIT &
$.848 \pm .008$ &
$.582 \pm .012$ &
$.266$ &
$.497 \pm .013$ &
$.902 \pm .011$ &
$.592 \pm .014$ &
$.562 \pm .014$ &
$.543 \pm .015$ \\

INC $\rightarrow$ INC &
$.845 \pm .033$ &
$.383 \pm .014$ &
$.461$ &
$.324 \pm .017$ &
$.833 \pm .024$ &
$.367 \pm .022$ &
$.553 \pm .050$ &
$.503 \pm .062$ \\

SVDB $\rightarrow$ SVDB &
$.830 \pm .014$ &
$.646 \pm .022$ &
$.184$ &
$.565 \pm .023$ &
$.846 \pm .017$ &
$.637 \pm .017$ &
$.405 \pm .029$ &
$.391 \pm .031$ \\

\midrule

MIT $\rightarrow$ INC &
$.874 \pm .022$ &
$.367 \pm .007$ &
$.507$ &
$.325 \pm .012$ &
$.815 \pm .029$ &
$.349 \pm .020$ &
$.559 \pm .051$ &
$.532 \pm .058$ \\

INC $\rightarrow$ MIT &
$.867 \pm .006$ &
$.551 \pm .012$ &
$.316$ &
$.485 \pm .013$ &
$.886 \pm .010$ &
$.584 \pm .013$ &
$.576 \pm .006$ &
$.566 \pm .009$ \\

MIT $\rightarrow$ SVDB &
$.876 \pm .015$ &
$.556 \pm .037$ &
$.320$ &
$.562 \pm .035$ &
$.866 \pm .026$ &
$.561 \pm .054$ &
$.392 \pm .052$ &
$.388 \pm .050$ \\

INC $\rightarrow$ SVDB &
$.897 \pm .008$ &
$.589 \pm .008$ &
$.308$ &
$.608 \pm .016$ &
$.859 \pm .019$ &
$.634 \pm .012$ &
$.475 \pm .025$ &
$.467 \pm .023$ \\

\bottomrule
\end{tabular}%
}

\vspace{0.3em}
\parbox{0.99\textwidth}{%
\scriptsize
\textit{Note:} MIT: MIT-BIH; INC: INCART; Acc.: accuracy;
BA: balanced accuracy; A--BA gap: difference between mean accuracy and mean
balanced accuracy; F1: macro-F1; AUROC: macro-AUROC;
AUPRC: macro-AUPRC; MCC: Matthews correlation coefficient;
$\kappa$: Cohen's kappa. Acc., BA, F1, AUROC, AUPRC, MCC, and $\kappa$
are reported as mean $\pm$ SD across five runs.
}

\end{table*}

Hybrid SSL showed measurable performance across all seven source--target routes under the patient-disjoint evaluation. The highest overall accuracy was obtained for INCART~$\rightarrow$~SVDB ($0.8968 \pm 0.0083$), followed by MIT-BIH~$\rightarrow$~SVDB ($0.8755 \pm 0.0147$) and MIT-BIH~$\rightarrow$~INCART ($0.8737 \pm 0.0222$). However, accuracy remained substantially higher than balanced accuracy in the INCART-target settings. MIT-BIH~$\rightarrow$~INCART achieved balanced accuracy of only $0.3669 \pm 0.0073$, while INCART~$\rightarrow$~INCART achieved $0.3834 \pm 0.0142$, indicating a strong influence of the majority classes on overall accuracy. The effect of majority classes has been discussed further in~\ref{subsec: class-imbal}.

The highest within-target balanced accuracy was obtained for SVDB~$\rightarrow$~SVDB at $0.6464 \pm 0.0217$, with macro-F1 of $0.5647 \pm 0.0232$. Among the cross-dataset routes, INCART~$\rightarrow$~SVDB achieved the highest balanced accuracy ($0.5889 \pm 0.0079$) and macro-F1 ($0.6077 \pm 0.0156$). INCART~$\rightarrow$~MIT-BIH achieved the highest cross-dataset macro-AUROC at $0.8861 \pm 0.0102$, whereas INCART~$\rightarrow$~SVDB achieved the highest cross-dataset macro-AUPRC at $0.6337 \pm 0.0116$. Absolute SVDB metric levels should not be compared directly with the five-class MIT-BIH and INCART routes as if task difficulty were identical, because SVDB was evaluated only on the shared N/S/V classes.

The two MIT-BIH--INCART cross-dataset routes produced different absolute performance profiles, but these values should not be interpreted as a direct comparison of transfer direction because the target dataset differs between the two experiments. Their transfer behaviour is therefore assessed relative to the corresponding within-target reference in Section~\ref{sec:generalisation_gaps}.

Overall, the Hybrid SSL encoder retained transferable representation structure across all seven evaluated routes, but class-balanced performance remained strongly dependent on the source--target combination and target class distribution. Ranking-based metrics remained comparatively stronger than balanced accuracy in several settings, while the low balanced accuracy and macro-F1 observed for the INCART-target evaluations indicate persistent difficulty in recognising minority classes under patient-disjoint testing.

\subsection{Architecture-Matched Baseline and Ablation Comparison}

Table~\ref{tab:matched_transformer_comparison} compares Hybrid SSL with architecture-matched reconstruction-only SSL, contrastive-only SSL, source-supervised Transformer pretraining, and a random frozen Transformer under the same frozen target linear-probing protocol. All conditions use the same Transformer encoder architecture, while their trainable parameter counts are summarised in Table~\ref{tab:matched_parameter_counts}. Differences between
the architecture-matched conditions can therefore be interpreted primarily in terms of the representation-learning objective rather than model-family differences.

Hybrid SSL produced higher mean balanced accuracy and macro-F1 than the random frozen Transformer across all seven source--target routes, confirming that the pretrained encoder learned useful representations beyond architectural inductive bias alone. Hybrid SSL produced higher mean balanced accuracy than the architecture-matched source-supervised Transformer in all seven routes. For macro-F1, Hybrid SSL was higher in four routes, whereas the source-supervised Transformer was slightly higher in three. The comparison should therefore be interpreted as showing a consistent advantage for Hybrid SSL in class-balanced recall, but a mixed route-dependent pattern for macro-F1, rather than a general superiority of self-supervised over supervised learning.

The reconstruction-only comparison showed that the benefit of adding the contrastive term was metric- and route-dependent. Hybrid SSL achieved higher balanced accuracy in four of the seven routes, whereas reconstruction-only SSL was higher in three. For macro-F1, Hybrid SSL was higher in five routes and reconstruction-only SSL in two. These findings indicate that the hybrid objective did not uniformly improve performance over masked reconstruction alone, although it showed a modest overall advantage across the evaluated routes.

In contrast, the contrastive-only Transformer performed substantially worse than both Hybrid SSL and reconstruction-only SSL across all seven routes. Balanced accuracy ranged from 0.2015 to 0.3520 for the contrastive-only condition, compared with 0.3669--0.6464 for Hybrid SSL. This indicates that, under the present augmentation and optimisation configuration, contrastive learning alone was insufficient to learn comparably useful heartbeat representations. The stronger performance of Hybrid SSL relative to contrastive-only SSL therefore appears to depend critically on the reconstruction component, while the incremental benefit of adding contrastive learning to reconstruction remains source--target dependent.

The architecture-matched source-supervised Transformer provides the most direct supervised source-pretraining comparison. Hybrid SSL achieved higher mean balanced accuracy than the source-supervised Transformer in all seven routes, including 0.5818 versus 0.5597 for MIT-BIH~$\rightarrow$~MIT-BIH, 0.3834 versus 0.3691 for INCART~$\rightarrow$~INCART, and 0.6464 versus 0.5668 for SVDB~$\rightarrow$~SVDB. Macro-F1 showed a less uniform pattern: Hybrid SSL was higher in four of the seven routes, while the source-supervised Transformer was slightly higher for INCART~$\rightarrow$~INCART, SVDB~$\rightarrow$~SVDB, and MIT-BIH~$\rightarrow$~INCART. These results do not establish that self-supervised learning is universally preferable to supervised learning; they show that, with the matched Transformer architecture, patient-disjoint data partitions, training budget, and downstream probe used here, Hybrid SSL produced consistently stronger balanced accuracy while macro-F1 remained route-dependent.

Table~\ref{tab:target_supervised_transformer} provides an additional target-supervised end-to-end reference. The end-to-end Transformer achieved balanced accuracies of $0.5502 \pm 0.0299$, $0.3607 \pm 0.0110$, and $0.5508 \pm 0.0310$ on MIT-BIH, INCART, and SVDB, respectively. These values were lower than the corresponding within-target Hybrid SSL frozen-representation results of $0.5818 \pm 0.0117$, $0.3834 \pm 0.0142$, and $0.6464 \pm 0.0217$. Mean macro-F1 showed the same ordering on all three datasets. However, the target-supervised condition addresses a different question from the frozen-transfer experiments because its encoder and classifier are jointly optimised using target labels. It should therefore be interpreted as a supervised classification reference rather than as a directly matched source-pretraining ablation. The results nevertheless show that, under the same patient-disjoint partitions and unweighted class-imbalanced training setting, end-to-end target supervision did not automatically yield stronger class-balanced performance than the representations obtained through self-supervised pretraining followed by linear probing.

\begin{table}[t]
\centering
\caption{Architecture-matched Transformer comparison under frozen target linear probing.}
\label{tab:matched_transformer_comparison}

\scriptsize
\setlength{\tabcolsep}{2.2pt}
\renewcommand{\arraystretch}{1.08}

\resizebox{.99\linewidth}{!}{%
\begin{tabular}{lccccc}
\toprule
\textbf{Train $\rightarrow$ Test} &
\textbf{Hyb.} &
\textbf{Rec.} &
\textbf{Con.} &
\textbf{Src.} &
\textbf{Rand.} \\
\midrule

\multicolumn{6}{l}{\textit{Balanced accuracy}} \\
\midrule

MIT $\rightarrow$ MIT &
$.582\!\pm\!.012$ &
\textbf{$.586\!\pm\!.018$} &
$.242\!\pm\!.029$ &
$.560\!\pm\!.029$ &
$.482\!\pm\!.031$ \\

INC $\rightarrow$ INC &
$.383\!\pm\!.014$ &
\textbf{$.387\!\pm\!.008$} &
$.221\!\pm\!.020$ &
$.369\!\pm\!.007$ &
$.281\!\pm\!.014$ \\

SVDB $\rightarrow$ SVDB &
\textbf{$.646\!\pm\!.022$} &
$.614\!\pm\!.024$ &
$.336\!\pm\!.005$ &
$.567\!\pm\!.060$ &
$.388\!\pm\!.018$ \\

MIT $\rightarrow$ INC &
$.367\!\pm\!.007$ &
\textbf{$.371\!\pm\!.015$} &
$.202\!\pm\!.003$ &
$.332\!\pm\!.025$ &
$.281\!\pm\!.014$ \\

INC $\rightarrow$ MIT &
\textbf{$.551\!\pm\!.012$} &
$.518\!\pm\!.023$ &
$.352\!\pm\!.077$ &
$.428\!\pm\!.071$ &
$.482\!\pm\!.031$ \\

MIT $\rightarrow$ SVDB &
\textbf{$.556\!\pm\!.037$} &
$.518\!\pm\!.025$ &
$.333\!\pm\!.002$ &
$.486\!\pm\!.045$ &
$.388\!\pm\!.018$ \\

INC $\rightarrow$ SVDB &
\textbf{$.589\!\pm\!.008$} &
$.587\!\pm\!.008$ &
$.346\!\pm\!.016$ &
$.429\!\pm\!.024$ &
$.388\!\pm\!.018$ \\

\midrule
\multicolumn{6}{l}{\textit{Macro-F1}} \\
\midrule

MIT $\rightarrow$ MIT &
\textbf{$.497\!\pm\!.013$} &
$.484\!\pm\!.024$ &
$.249\!\pm\!.036$ &
$.458\!\pm\!.014$ &
$.482\!\pm\!.035$ \\

INC $\rightarrow$ INC &
$.324\!\pm\!.017$ &
\textbf{$.335\!\pm\!.012$} &
$.225\!\pm\!.032$ &
$.325\!\pm\!.012$ &
$.304\!\pm\!.015$ \\

SVDB $\rightarrow$ SVDB &
$.565\!\pm\!.023$ &
$.554\!\pm\!.026$ &
$.318\!\pm\!.009$ &
\textbf{$.574\!\pm\!.057$} &
$.410\!\pm\!.026$ \\

MIT $\rightarrow$ INC &
$.325\!\pm\!.012$ &
\textbf{$.336\!\pm\!.023$} &
$.192\!\pm\!.006$ &
$.326\!\pm\!.026$ &
$.304\!\pm\!.015$ \\

INC $\rightarrow$ MIT &
\textbf{$.485\!\pm\!.013$} &
$.454\!\pm\!.028$ &
$.331\!\pm\!.062$ &
$.415\!\pm\!.061$ &
$.482\!\pm\!.035$ \\

MIT $\rightarrow$ SVDB &
\textbf{$.562\!\pm\!.035$} &
$.527\!\pm\!.022$ &
$.314\!\pm\!.001$ &
$.491\!\pm\!.044$ &
$.410\!\pm\!.026$ \\

INC $\rightarrow$ SVDB &
\textbf{$.608\!\pm\!.016$} &
$.599\!\pm\!.009$ &
$.333\!\pm\!.021$ &
$.447\!\pm\!.033$ &
$.410\!\pm\!.026$ \\

\bottomrule
\end{tabular}%
}

\vspace{0.3em}
\parbox{0.98\linewidth}{%
\scriptsize
\textit{Note:} MIT: MIT-BIH; INC: INCART; Hyb.: Hybrid SSL; Rec.: reconstruction-only SSL; Con.: contrastive-only SSL;
Src.: source-supervised Transformer; Rand.: randomly initialised frozen Transformer.
Bold values indicate the highest mean performance for each source--target setting.
}

\end{table}

\begin{table}[t]
\centering
\caption{Target-supervised performance of the end-to-end Transformer model.
Values are reported as mean $\pm$ standard deviation.}
\label{tab:target_supervised_transformer}

\small
\renewcommand{\arraystretch}{1.15}
\setlength{\tabcolsep}{4pt}
\resizebox{.99\linewidth}{!}{%
\begin{tabular}{lccc}
\toprule
\textbf{Metric} &
\textbf{MIT-BIH} &
\textbf{INCART} &
\textbf{SVDB} \\
\midrule

Accuracy
& $0.816 \pm 0.076$
& $0.840 \pm 0.028$
& $0.850 \pm 0.029$ \\

Balanced Accuracy
& $0.550 \pm 0.030$
& $0.361 \pm 0.011$
& $0.551 \pm 0.031$ \\

Macro-F1
& $0.459 \pm 0.030$
& $0.305 \pm 0.006$
& $0.538 \pm 0.044$ \\

Macro-AUROC
& $0.861 \pm 0.034$
& $0.774 \pm 0.043$
& $0.834 \pm 0.008$ \\

Macro-AUPRC
& $0.504 \pm 0.040$
& $0.314 \pm 0.026$
& $0.548 \pm 0.045$ \\

MCC
& $0.519 \pm 0.092$
& $0.519 \pm 0.045$
& $0.339 \pm 0.061$ \\

Cohen's $\kappa$
& $0.492 \pm 0.114$
& $0.474 \pm 0.052$
& $0.335 \pm 0.058$ \\

\bottomrule
\end{tabular}}
\end{table}

\begin{table}[t]
\centering
\caption{Trainable parameter counts across architecture-matched conditions.
All conditions use the same Transformer encoder with 3,358,592 parameters.}
\label{tab:matched_parameter_counts}
\small
\setlength{\tabcolsep}{5pt}
\renewcommand{\arraystretch}{1.12}

\begin{tabular}{lr}
\toprule
\textbf{Condition} & \textbf{Trainable parameters} \\
\midrule
Hybrid SSL                     & 3,482,176 \\
Reconstruction-only SSL        & 3,457,408 \\
Contrastive-only SSL           & 3,383,360 \\
\midrule
Source-supervised (5-class)    & 3,359,237 \\
Source-supervised (3-class)    & 3,358,979 \\
Target-supervised (5-class)    & 3,359,237 \\
Target-supervised (3-class)    & 3,358,979 \\
\midrule
Random frozen Transformer      & -- \\
\bottomrule
\end{tabular}

\vspace{0.4em}
\footnotesize
\textit{Note:} Parameter counts include all trainable components used during
pretraining or supervised training. The random Transformer is not pretrained;
its encoder remains frozen during downstream evaluation.
\end{table}

\subsection{Target-Normalised Transfer Gaps}
\label{sec:generalisation_gaps}

To compare cross-dataset generalisation across the principal representation-learning strategies, Table~\ref{tab:transfer_generalisation} reports both the absolute cross-dataset performance and the corresponding target-normalised transfer gap
for Hybrid SSL, reconstruction-only SSL, and the source-supervised Transformer. For each method, the gap compares cross-dataset performance with that method's own within-target reference on the same target dataset. Balanced accuracy and macro-F1 are reported because they are the principal class-balanced metrics used throughout this study.

\begin{table}[t]
\centering
\caption{Cross-dataset performance and target-normalised transfer gaps. Panel A compares balanced accuracy (BA) and Panel B compares macro-F1
across the three principal representation-learning strategies.}
\label{tab:transfer_generalisation}

\scriptsize
\renewcommand{\arraystretch}{1.10}
\setlength{\tabcolsep}{4pt}

\textbf{(a) Balanced accuracy and target-normalised gaps}

\vspace{0.3em}

\begin{tabular}{lccc}
\toprule
\textbf{Transfer} &
\textbf{Hyb. BA ($\Delta$BA)} &
\textbf{Rec. BA ($\Delta$BA)} &
\textbf{Src. BA ($\Delta$BA)} \\
\midrule

MIT $\rightarrow$ INC &
$.367\,(-.016)$ &
$\textbf{.371}\,(-.016)$ &
$.332\,(-.038)$ \\

INC $\rightarrow$ MIT &
$\textbf{.551}\,(-.031)$ &
$.518\,(-.068)$ &
$.428\,(-.132)$ \\

MIT $\rightarrow$ SVDB &
$\textbf{.556}\,(-.091)$ &
$.518\,(-.096)$ &
$.486\,(-.080)$ \\

INC $\rightarrow$ SVDB &
$\textbf{.589}\,(-.057)$ &
$.587\,(-.027)$ &
$.429\,(-.138)$ \\

\bottomrule
\end{tabular}

\vspace{0.8em}

\textbf{(b) Macro-F1 and target-normalised gaps}

\vspace{0.3em}

\begin{tabular}{lccc}
\toprule
\textbf{Transfer} &
\textbf{Hyb. F1 ($\Delta$F1)} &
\textbf{Rec. F1 ($\Delta$F1)} &
\textbf{Src. F1 ($\Delta$F1)} \\
\midrule

MIT $\rightarrow$ INC &
$.325\,(+.001)$ &
$\textbf{.336}\,(+.001)$ &
$.326\,(+.001)$ \\

INC $\rightarrow$ MIT &
$\textbf{.485}\,(-.012)$ &
$.454\,(-.030)$ &
$.415\,(-.043)$ \\

MIT $\rightarrow$ SVDB &
$\textbf{.562}\,(-.003)$ &
$.527\,(-.026)$ &
$.491\,(-.084)$ \\

INC $\rightarrow$ SVDB &
$\textbf{.608}\,(+.043)$ &
$.598\,(+.045)$ &
$.447\,(-.128)$ \\

\bottomrule
\end{tabular}

\vspace{0.4em}

\parbox{0.95\linewidth}{%
\scriptsize
\textit{Note:} MIT: MIT-BIH; INC: INCART; Hyb.: Hybrid SSL;
Rec.: reconstruction-only SSL; Src.: source-supervised Transformer.
Values report mean cross-dataset performance, with the corresponding
target-normalised gap in parentheses. For each method, $\Delta$ is defined as
cross-dataset performance minus that method's corresponding within-target
reference. Negative values indicate degradation relative to within-target
training, whereas positive values indicate improvement. Bold values indicate
the highest mean cross-dataset performance for each transfer route and metric.
}

\end{table}

The comparative results show that Hybrid SSL achieved the highest mean cross-dataset balanced accuracy and macro-F1 in three of the four transfer routes. For INCART $\rightarrow$ MIT-BIH, Hybrid SSL achieved a balanced accuracy of 0.551 and macro-F1 of 0.485, compared with 0.518 and 0.454 for reconstruction-only SSL and 0.428 and 0.415 for the source-supervised Transformer. Hybrid SSL was also strongest for both transfers to SVDB, achieving balanced accuracy of 0.556 and 0.589 for MIT-BIH $\rightarrow$ SVDB and INCART $\rightarrow$ SVDB, respectively. Reconstruction-only SSL produced the highest values for MIT-BIH $\rightarrow$ INCART, although its advantage over Hybrid SSL was small for both balanced accuracy (0.371 versus 0.367) and macro-F1 (0.336 versus 0.325).

The target-normalised gaps provide complementary information about how much each method changed relative to its own within-target reference. These gaps did not uniformly favour Hybrid SSL. For example, the source-supervised Transformer had the smallest balanced-accuracy reduction for MIT-BIH$\rightarrow$ SVDB, despite having substantially lower absolute cross-dataset balanced accuracy than Hybrid SSL (0.486 versus 0.556). Similarly, reconstruction-only SSL had the smallest balanced-accuracy reduction for INCART $\rightarrow$ SVDB, while Hybrid SSL achieved the higher absolute balanced accuracy and macro-F1. Thus, resistance to degradation relative to within-target performance was method and route-dependent, whereas Hybrid SSL provided the strongest absolute cross-dataset performance in three of the four evaluated transfer settings.

\subsection{Class-Imbalance Effects}
\label{subsec: class-imbal}
Class imbalance considerably affected how the transfer results should be interpreted, particularly for INCART-target evaluation. The patient-disjoint target partitions were strongly imbalanced (Table~\ref{tab:classaware_splits}); an N-only classifier would achieve accuracies of 0.852, 0.886, and 0.880 on the MIT-BIH, INCART, and SVDB test sets, respectively, despite balanced accuracy remaining at the corresponding chance level. Notably, the INCART majority-class reference accuracy of 0.886 exceeds the mean raw accuracy of both INCART~$\rightarrow$~INCART (0.845) and MIT-BIH~$\rightarrow$~INCART (0.874). Raw accuracy alone therefore does not establish whether a learned representation retains useful class-discriminative information.

Hybrid SSL nevertheless achieved balanced accuracy substantially above chance in all evaluated settings, showing that its representations did not simply collapse to majority-class prediction. Table~\ref{tab:hybrid_overall_transfer} also quantifies the divergence between accuracy and balanced accuracy. The two largest gaps occurred in the INCART-target settings, indicating that useful representation structure was retained but did not translate into uniformly strong recall across the five AAMI classes. 

\begin{figure*}[t]
\centering
\includegraphics[width=0.85\textwidth]{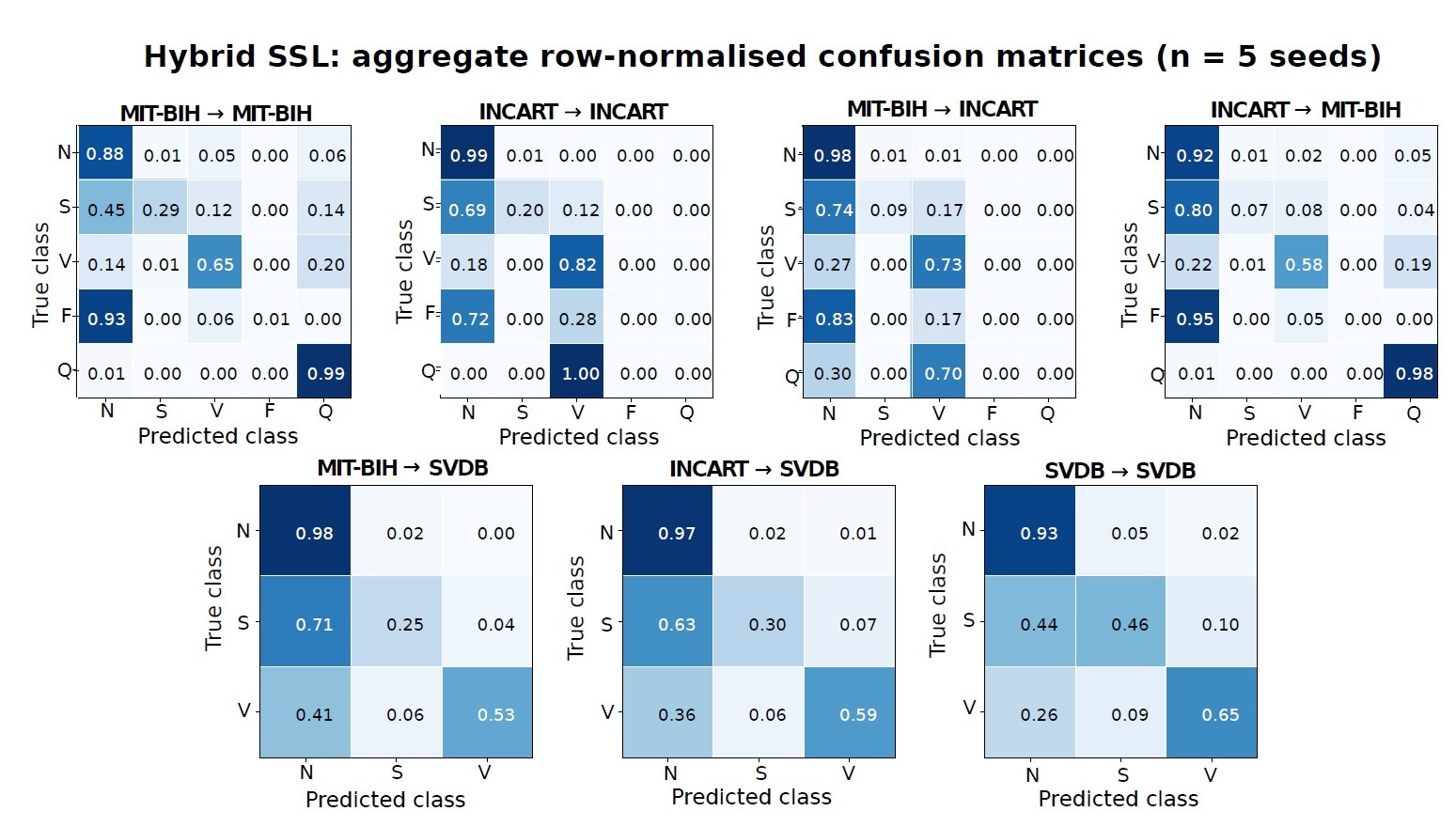}
\caption{Aggregated row-normalised confusion matrices for Hybrid SSL across the seven source--target routes. }
\label{fig:hybrid_ssl_confusion_matrices}
\end{figure*}

The row-normalised confusion matrices in Fig.~\ref{fig:hybrid_ssl_confusion_matrices} further show that the performance limitation was class-specific rather than a uniform degradation across heartbeat categories.

Because INCART contains only two Q-class test beats, a sensitivity analysis was performed to assess how strongly this extremely sparse class affected the aggregate class-balanced metrics. Without retraining the models or changing their predictions, balanced accuracy and macro-F1 were recalculated for each seeded run after excluding Q from metric aggregation. Table~\ref{tab:incart_q_sensitivity} shows substantial improvements in both metrics; however, balanced accuracy remained below 0.50 for both INCART~$\rightarrow$~INCART and MIT-BIH~$\rightarrow$~INCART. The poor class-balanced performance therefore cannot be attributed solely to the two-example Q class. The original five-class results remain the principal analysis for consistency with the predefined AAMI mapping, while Q-dependent estimates should be interpreted cautiously.

\begin{table}[t]
\centering
\caption{Effect of excluding the sparsely represented Q class from 
INCART-target evaluation. Values are mean $\pm$ SD across five runs.}
\label{tab:incart_q_sensitivity}
\scriptsize
\setlength{\tabcolsep}{4pt}
\renewcommand{\arraystretch}{1}
\resizebox{\columnwidth}{!}{%
\begin{tabular}{lccc}
\toprule
\textbf{Experiment} & \textbf{Classes} & \textbf{BA} & \textbf{Macro-F1} \\
\midrule

INCART $\rightarrow$ INCART
& 5-class
& $0.3834 \pm 0.0142$
& $0.3238 \pm 0.0169$ \\

& Excl.\ Q
& $\mathbf{0.4792 \pm 0.0177}$
& $\mathbf{0.4048 \pm 0.0211}$ \\

\midrule

MIT-BIH $\rightarrow$ INCART
& 5-class
& $0.3669 \pm 0.0073$
& $0.3248 \pm 0.0124$ \\

& Excl.\ Q
& $\mathbf{0.4586 \pm 0.0091}$
& $\mathbf{0.4060 \pm 0.0155}$ \\

\bottomrule
\end{tabular}
}
\vspace{-4mm}
\end{table}

The classwise results in Table~\ref{tab:incart_classwise} clarify the remaining limitation. Q achieved zero recall, PPV, and F1 in both INCART-target settings, and its ranking metrics are inherently unstable given only two positive test examples. More importantly, poor recognition was not confined to Q: F- and S-class recall also remained very low, whereas N and V were recognised substantially more reliably. This pattern occurred for both MIT-BIH~$\rightarrow$~INCART and the INCART~$\rightarrow$~INCART within-target reference. The minority-class weakness is therefore not attributable solely to cross-dataset transfer; it also reflects the severe class distribution and patient-disjoint characteristics of the INCART target.

\begin{table}[t]
\centering
\caption{Classwise Hybrid SSL performance for the two INCART-target experiments.}
\label{tab:incart_classwise}

\scriptsize
\setlength{\tabcolsep}{2.5pt}
\renewcommand{\arraystretch}{1.10}

\resizebox{0.99\linewidth}{!}{%
\begin{tabular}{clrrrrrr}
\toprule
\textbf{Exp.} &
\textbf{C} &
\textbf{Sup.} &
\textbf{Rec.} &
\textbf{PPV} &
\textbf{F1} &
\textbf{AUROC} &
\textbf{AUPRC} \\
\midrule

\multirow{5}{*}{%
\rotatebox[origin=c]{90}{\textbf{INC $\rightarrow$ INC}}}
& N & 49,581
& $.841 \pm .036$
& $.986 \pm .001$
& $.907 \pm .022$
& $.939 \pm .009$
& $.990 \pm .001$ \\

& S & 625
& $.104 \pm .049$
& $.041 \pm .040$
& $.055 \pm .047$
& $.602 \pm .028$
& $.031 \pm .024$ \\

& V & 5,655
& $.973 \pm .005$
& $.497 \pm .044$
& $.657 \pm .037$
& $.974 \pm .012$
& $.807 \pm .103$ \\

& F & 72
& $.000 \pm .000$
& $.000 \pm .000$
& $.000 \pm .000$
& $.731 \pm .031$
& $.004 \pm .002$ \\

& Q & 2
& $.000 \pm .000$
& $.000 \pm .000$
& $.000 \pm .000$
& $.917 \pm .085$
& $.005 \pm .004$ \\

\midrule

\multirow{5}{*}{%
\rotatebox[origin=c]{90}{\textbf{MIT $\rightarrow$ INC}}}
& N & 49,581
& $.885 \pm .024$
& $.975 \pm .003$
& $.928 \pm .014$
& $.924 \pm .019$
& $.987 \pm .003$ \\

& S & 625
& $.072 \pm .044$
& $.073 \pm .038$
& $.070 \pm .036$
& $.670 \pm .044$
& $.039 \pm .024$ \\

& V & 5,655
& $.875 \pm .024$
& $.483 \pm .055$
& $.621 \pm .049$
& $.954 \pm .015$
& $.712 \pm .090$ \\

& F & 72
& $.003 \pm .006$
& $.067 \pm .149$
& $.005 \pm .012$
& $.703 \pm .041$
& $.006 \pm .003$ \\

& Q & 2
& $.000 \pm .000$
& $.000 \pm .000$
& $.000 \pm .000$
& $.823 \pm .076$
& $.001 \pm .001$ \\

\bottomrule
\end{tabular}%
}

\vspace{0.3em}
\parbox{0.95\linewidth}{%
\scriptsize
\textit{Note:} MIT: MIT-BIH; INC: INCART; Exp.: experiment; 
C: class; Sup.: support; Rec.: recall; PPV: positive predictive value;
AUROC: area under the receiver operating characteristic curve; 
AUPRC: area under the precision--recall curve.
Values are mean $\pm$ SD across five runs.
}
\vspace{-4mm}
\end{table}

Overall, these results distinguish useful representation transfer from balanced arrhythmia recognition. Hybrid SSL retained class-discriminative information beyond a trivial majority-class solution, but neither cross-dataset transfer nor target-specific self-supervised pretraining produced reliable recognition of all minority classes under the INCART patient-disjoint evaluation. The Q-exclusion analysis further demonstrates that extreme Q sparsity explains only part of the observed class-balanced performance limitation.

The discrepancy between accuracy and balanced accuracy was less pronounced for the MIT-BIH and SVDB targets than for INCART. However, SVDB was evaluated only on the N/S/V classes and excluded the severely under-represented F and Q categories; its class-balanced results are therefore not directly comparable with the five-class MIT-BIH and INCART settings.

\section{Discussion}
\label{sec:disc}

\begin{table*}[t]
\centering
\caption{Summary of the principal experimental findings.}
\label{tab:key_findings}
\scriptsize
\renewcommand{\arraystretch}{1.15}
\begin{tabularx}{\linewidth}{@{}p{3.5cm}p{2.4cm}X@{}}
\toprule
\textbf{Comparison} & \textbf{Outcome} & \textbf{Evidence} \\
\midrule

Hybrid SSL vs.\ Random Transformer
& \textbf{Hybrid SSL}
& Higher balanced accuracy and macro-F1 in all seven source--target evaluation routes. \\

Hybrid SSL vs.\ Source-supervised Transformer
& \textbf{Hybrid SSL for BA}
& Higher balanced accuracy in all seven routes. Macro-F1 was higher for Hybrid SSL in 4/7 routes and for the source-supervised Transformer in 3/7 routes. \\

Hybrid SSL vs.\ Reconstruction-only SSL
& \textbf{Mixed}
& Hybrid SSL achieved higher balanced accuracy in 4/7 routes and higher macro-F1 in 5/7 routes, confirming that the incremental effect of the contrastive objective was route- and metric-dependent. \\

Hybrid SSL vs.\ Contrastive-only SSL
& \textbf{Hybrid SSL}
& Contrastive-only SSL produced substantially lower balanced accuracy and macro-F1 across all seven evaluation routes. \\

Cross-dataset generalisation
& \textbf{Hybrid SSL in 3/4 routes}
& Hybrid SSL achieved the highest cross-dataset balanced accuracy and macro-F1
in three of four transfer routes; reconstruction-only SSL was highest for
MIT-BIH $\rightarrow$ INCART. Target-normalised degradation remained
method and route-dependent. \\

Minority-class recognition
& \textbf{Unresolved}
& Near-zero fusion-beat (F) recall and consistently low supraventricular ectopic-beat (S) recall in several five-class evaluations. \\

\bottomrule
\end{tabularx}
\end{table*}

Table~\ref{tab:key_findings} summarises the principal experimental findings. This study evaluated whether self-supervised ECG representations learned from one arrhythmia database can transfer to independent target databases under patient-disjoint evaluation. The proposed Hybrid SSL encoder consistently outperformed the random frozen Transformer across all seven source--target routes, showing that source-domain self-supervised pretraining learned transferable ECG structure beyond architectural inductive bias alone. The architecture-matched ablations further clarify where this benefit originated. Reconstruction-only SSL remained highly competitive with Hybrid SSL, with Hybrid SSL achieving higher balanced accuracy in four of seven routes and higher macro-F1 in five of seven routes. In contrast, contrastive-only SSL performed substantially worse than both reconstruction-only and Hybrid SSL across all routes. These findings indicate that masked reconstruction was a critical contributor to useful heartbeat representations under the tested configuration, whereas the additional contrastive objective provided a modest and source--target-dependent benefit rather than a uniform improvement.

The architecture-matched source-supervised Transformer provides a controlled comparison of source supervision because it used the same encoder architecture, source partitions, training budget, and frozen target-probing protocol. Hybrid SSL achieved higher mean balanced accuracy in all seven routes, while macro-F1 was higher for Hybrid SSL in four routes and for the source-supervised Transformer in three. This should not be interpreted as evidence that self-supervised learning is generally superior to supervised ECG learning; the comparison is specific to the present architecture, datasets, optimisation settings, class distributions, and frozen linear-probe evaluation. The target-supervised end-to-end Transformer also addresses a different experimental question because both its encoder and classifier were optimised directly using target labels. Its inclusion therefore provides an additional supervised classification reference rather than a matched source-pretraining ablation. Together, these comparisons show that the pretraining objective materially affected frozen representation quality while highlighting the importance of architecture- and task-matched baselines when attributing performance differences to self-supervision.

The results also demonstrate the importance of class-sensitive evaluation. Raw accuracy substantially overstated performance in the most imbalanced settings, particularly when INCART was the target dataset. For example, MIT-BIH~$\rightarrow$~INCART achieved an accuracy of 0.8737 but a balanced accuracy of only 0.3669, while INCART~$\rightarrow$~INCART achieved 0.8445 and 0.3834, respectively. Classwise analysis showed that this divergence was driven by weak minority-class recognition: Q recall was zero in both INCART-target settings, F recall was zero or near zero, and S recall remained low, whereas N and V were comparatively better recognised. The Q-excluded sensitivity analysis further showed that the extremely small INCART Q test set materially depressed the five-class balanced accuracy and macro-F1 values. These findings reinforce that high overall accuracy cannot be interpreted as uniformly reliable performance across AAMI classes and justify the use of balanced accuracy, macro-F1, classwise recall, ranking-based metrics, and confusion-matrix analysis alongside raw accuracy.

The comparative cross-dataset analysis further clarifies the generalisation behaviour of the learned representations. Hybrid SSL achieved the highest mean cross-dataset balanced accuracy and macro-F1 in three of the four transfer routes, while reconstruction-only SSL was strongest for MIT-BIH~$\rightarrow$~INCART. In particular, Hybrid SSL outperformed both reconstruction-only SSL and the source-supervised Transformer on both metrics for INCART~$\rightarrow$~MIT-BIH, MIT- BIH~$\rightarrow$~SVDB, and INCART~$\rightarrow$~SVDB. This advantage was not equivalent to consistently having the smallest target-normalised degradation, however. The relative gaps were method- and route-dependent; for example, the source-supervised Transformer showed a smaller balanced-accuracy reduction than Hybrid SSL for
MIT-BIH~$\rightarrow$~SVDB despite substantially lower absolute cross-dataset balanced accuracy. Thus, the results support stronger absolute cross-dataset performance for Hybrid SSL in most evaluated transfer routes, while also showing that resistance to degradation relative to within-target training is not uniformly superior. Useful representation transfer therefore does not imply that domain shift has been eliminated.

Several limitations should be considered. First, although Hybrid SSL source pretraining was unlabelled, labelled target training records were still required to fit the linear probe; the framework should therefore be interpreted as label-efficient source pretraining rather than fully label-free arrhythmia classification. Second, the experiments used public ECG databases and a single-lead setting, and further validation is required on larger, prospective, multi-centre, and multi-lead datasets before clinical generalisability can be assessed. The results should therefore be interpreted as methodological evidence concerning ECG representation transfer rather than evidence of diagnostic utility, patient-level effectiveness, or deployment readiness. Third, the use of a linear probe was intentional because it isolates the quality of the frozen representation, but it does not establish whether non-linear target classifiers, calibration methods, cost-sensitive learning, or class-balanced decision rules could improve minority-class operating points.  Finally, the extremely small support of some minority classes limits the stability of class-specific estimates. In particular, INCART contained only four Q training beats and two Q test beats, while the F test set contained 72 beats. The corresponding Q-specific recall, precision, F1, AUROC, and AUPRC values should therefore be interpreted descriptively rather than as population-level estimates. Patient-disjoint splitting prevents the same patient from contributing to both training and test partitions, but it does not remove broader sources of domain shift arising from differences in annotation policy, acquisition conditions, and class prevalence across datasets.

Future work should investigate semi-supervised and few-shot target adaptation, class-balanced calibration, multi-lead representation learning, and domain-aware pretraining. It would also be valuable to determine whether the ranking structure learned by Hybrid SSL can be converted into more reliable minority-class decisions through calibrated thresholds, cost-sensitive target probing, or other target-efficient adaptation strategies without compromising patient-disjoint evaluation.


\section{Conclusion}
\label{conc}

This study presented TransECG-SSL, a hybrid self-supervised framework for learning transferable ECG heartbeat representations under patient-disjoint evaluation. Across the four cross-dataset transfer routes, Hybrid SSL achieved the highest mean balanced accuracy and macro-F1 in three settings, while reconstruction-only SSL was strongest for MIT-BIH $\rightarrow$ INCART. Hybrid SSL also consistently outperformed the random frozen Transformer and showed stronger cross-dataset performance than the architecture-matched source-supervised Transformer in the evaluated transfer settings. Reconstruction-based learning contributed strongly to representation quality, whereas the incremental benefit of the contrastive objective remained source--target dependent.

Despite this transferability, Hybrid SSL did not consistently show the smallest degradation relative to its own within-target reference, and its balanced accuracy remained below target-specific Hybrid SSL in all four cross-dataset routes. This distinction indicates that strong absolute cross-dataset performance does not imply complete invariance to domain shift. Minority-class recognition also remained challenging, particularly in the five-class settings. Overall, TransECG-SSL provides a label-efficient approach to cross-dataset ECG representation learning and demonstrates strong comparative transfer performance, while motivating further work on domain-robust pretraining and improved minority-class representation.

\section*{Code availability}

The source code supporting this study is maintained at \url{https://github.com/ahtisham-86/TransECG-SSL} and will be made publicly available after publication.

\section*{References}

\bibliographystyle{IEEEtran}
\bibliography{reference}

@article{hu2025novel,
  title={A novel multimodal self-supervised framework for ECG arrhythmia classification},
  author={Hu, Jianqiang and Li, Cheng and Cao, Jinde and Kou, Bo},
  journal={Computers in Biology and Medicine},
  volume={198},
  pages={111137},
  year={2025},
  note    = {{DOI}: \doi{10.1016/j.compbiomed.2025.111137}}
}

@article{zhang2025sequential,
  title={A sequential MAE-clustering self-supervised learning method for arrhythmia detection},
  author={Zhang, Yatao and Li, Xiangyu and Zhang, Liting and Wang, Jilin and Jiang, Shipeng and Ma, Yijun and Li, Dong},
  journal={Expert Systems with Applications},
  volume={269},
  pages={126379},
  year={2025},
  publisher={Elsevier},
  note    = {{DOI}: \doi{10.1016/j.eswa.2025.126379}}
}

@article{dias2021arrhythmia,
  title={Arrhythmia classification from single-lead ECG signals using the inter-patient paradigm},
  author={Dias, Felipe Meneguitti and Monteiro, Henrique LM and Cabral, Thales Wulfert and Naji, Rayen and Kuehni, Michael and Luz, Eduardo Jose da S},
  journal={Computer Methods and Programs in Biomedicine},
  volume={202},
  pages={105948},
  year={2021},
  publisher={Elsevier},
  note    = {{DOI}: \doi{10.1016/j.cmpb.2021.105948}}
}

@article{lyu2023automated,
  title={Automated inter-patient arrhythmia classification with dual attention neural network},
  author={Lyu, He and Li, Xiangkui and Zhang, Jian and Zhou, Chenchen and Tang, Xuezhi and Xu, Fanxin and Yang, Ye and Huang, Qinzhen and Xiang, Wei and Li, Dong},
  journal={Computer Methods and Programs in Biomedicine},
  volume={236},
  pages={107560},
  year={2023},
  publisher={Elsevier},
  note    = {{DOI}: \doi{10.1016/j.cmpb.2023.107560}}
}

@article{chen2021clecg,
  title={CLECG: A novel contrastive learning framework for electrocardiogram arrhythmia classification},
  author={Chen, Hui and Wang, Guijin and Zhang, Guodong and Zhang, Ping and Yang, Huazhong},
  journal={IEEE Signal Processing Letters},
  volume={28},
  pages={1993--1997},
  year={2021},
  publisher={IEEE},
  note    = {{DOI}: \doi{10.1109/LSP.2021.3114119}}
}

@article{xiao2023deep,
  title={Deep learning-based ECG arrhythmia classification: A systematic review},
  author={Xiao, Qiao and Lee, Khuan and Mokhtar, Siti Aisah and Ismail, Iskasymar and Pauzi, Ahmad Luqman bin Md and Zhang, Qiuxia and Lim, Poh Ying},
  journal={Applied Sciences},
  volume={13},
  number={8},
  pages={4964},
  year={2023},
  publisher={MDPI},
  note    = {{DOI}: \doi{10.3390/app13084964}}
}

@article{maghsoodi2026domain,
  title={Domain Knowledge is Power: Leveraging Physiological Priors for Self-Supervised Representation Learning in Electrocardiography},
  author={Maghsoodi, Nooshin and Nassar, Sarah and Wilson, Paul FR and To, Minh Nguyen Nhat and Mannina, Sophia and Addas, Shamel and Sibley, Stephanie and Pichora, David and Maslove, David and Abolmaesumi, Purang and others},
  journal={IEEE Transactions on Biomedical Engineering},
  year={2026},
  publisher={IEEE},
  note    = {{DOI}: \doi{10.1109/TBME.2026.3656904}}
}

@article{mavaddati2025ecg,
  title={ECG arrhythmias classification based on deep learning methods and transfer learning technique},
  author={Mavaddati, Samira},
  journal={Biomedical Signal Processing and Control},
  volume={101},
  pages={107236},
  year={2025},
  publisher={Elsevier},
  note    = {{DOI}: \doi{10.1016/j.bspc.2024.107236}}
}

@article{kim2026enhancing,
  title={Enhancing ECG Classification Generalization Through Unified Multi-Dataset Training},
  author={Kim, Minchan and Shin, Miyoung},
  journal={Sensors},
  volume={26},
  number={6},
  pages={1830},
  year={2026},
  publisher={MDPI},
  note    = {{DOI}: \doi{10.3390/s26061830}}
}

@article{han2022hctnet,
  title={HCTNet: An experience-guided deep learning network for inter-patient arrhythmia classification on imbalanced dataset},
  author={Han, Chuanqi and Wang, Peng and Huang, Ruoran and Cui, Li},
  journal={Biomedical Signal Processing and Control},
  volume={78},
  pages={103910},
  year={2022},
  publisher={Elsevier},
  note    = {{DOI}: \doi{10.1016/j.bspc.2022.103910}}
}

@article{rafi2022heartnet,
  title={HeartNet: Self multihead attention mechanism via convolutional network with adversarial data synthesis for ECG-based arrhythmia classification},
  author={Rafi, Taki Hasan and Ko, Young Woong},
  journal={IEEE Access},
  volume={10},
  pages={100501--100512},
  year={2022},
  publisher={IEEE},
  note    = {{DOI}: \doi{10.1109/ACCESS.2022.3206431}}
}

@article{li2022inter,
  title={Inter-patient arrhythmia classification with improved deep residual convolutional neural network},
  author={Li, Yuanlu and Qian, Renfei and Li, Kun},
  journal={Computer Methods and Programs in Biomedicine},
  volume={214},
  pages={106582},
  year={2022},
  publisher={Elsevier},
  note    = {{DOI}: \doi{10.1016/j.cmpb.2021.106582}}
}

@article{xu2023inter,
  title={Inter-patient ECG classification with intra-class coherence based weighted kernel extreme learning machine},
  author={Xu, Yuefan and Zhang, Sen and Xiao, Wendong},
  journal={Expert Systems with Applications},
  volume={227},
  pages={120095},
  year={2023},
  publisher={Elsevier},
  note    = {{DOI}: \doi{10.1016/j.eswa.2023.120095}}
}

@article{niu2019inter,
  title={Inter-patient ECG classification with symbolic representations and multi-perspective convolutional neural networks},
  author={Niu, Jinghao and Tang, Yongqiang and Sun, Zhengya and Zhang, Wensheng},
  journal={IEEE journal of biomedical and health informatics},
  volume={24},
  number={5},
  pages={1321--1332},
  year={2019},
  publisher={IEEE},
  note    = {{DOI}: \doi{10.1109/JBHI.2019.2942938}}
}

@misc{niloy2025lead,
  title={Lead-Aware Multi-Resolution Transformer With Domain Adaptation for Beat-Level ECG Arrhythmia Classification, Vol. 6 (2025), IEEE Open Journal of the Computer Society, ISSN: 2644-1268},
  author={Niloy, M and Islam, MT and Ullah, MS and Alom, J and Ahmed, M and Mridha, MF and Hossen, MJ},
  year={2025},
  note    = {{DOI}: \doi{10.1109/OJCS.2025.3637851}}
}

@article{zhang2022maefe,
  title={Maefe: Masked autoencoders family of electrocardiogram for self-supervised pretraining and transfer learning},
  author={Zhang, Huaicheng and Liu, Wenhan and Shi, Jiguang and Chang, Sheng and Wang, Hao and He, Jin and Huang, Qijun},
  journal={IEEE Transactions on Instrumentation and Measurement},
  volume={72},
  pages={1--15},
  year={2022},
  publisher={IEEE},
  note    = {{DOI}: \doi{10.1109/TIM.2022.3228267}}
}

@article{zakaria2024morphological,
  title={Morphological arrhythmia classification based on inter-patient and two leads ECG using machine learning},
  author={Zakaria, Hasballah and Nurdiniyah, Elsa Sari Hayunah and Kurniawati, Astri Maria and Naufal, Dziban and Sutisna, Nana},
  journal={IEEE Access},
  volume={12},
  pages={147372--147386},
  year={2024},
  publisher={IEEE},
  note    = {{DOI}: \doi{10.1109/ACCESS.2024.3469640}}
}

@article{wu2025one,
  title={One model, dual tasks: a novel distributionally adaptive learning framework for ecg classification and generation addressing intra-and inter-patient variability},
  author={Wu, Xinyi and Rezki, Zouheir and Balusu, Bhavan and Zhou, Jiaming and Bai, Jun},
  journal={Expert Systems with Applications},
  pages={130096},
  year={2025},
  publisher={Elsevier},
  note    = {{DOI}: \doi{10.1016/j.eswa.2025.130096}}
}

@article{dehghani2026rare,
  title={Rare-Beat Detection in Heterogeneous ECG: Federated Learning Under Severe Class Imbalance and Class-Missing Clients},
  author={Dehghani, Somayyeh and Pierre, Samuel},
  journal={IEEE Access},
  year={2026},
  publisher={IEEE},
  note    = {{DOI}: \doi{10.1109/ACCESS.2026.3715579}}
}

@article{li2025research,
  title={Research on cross-dataset cardiac signal domain generalization and feature interpretability},
  author={Li, Runwen and Aierken, Yierpani and Xu, Yu and Liu, Jiang and Tang, Yongjiang},
  journal={Scientific Reports},
  volume={16},
  number={1},
  pages={3138},
  year={2025},
  publisher={Nature Publishing Group UK London},
  note    = {{DOI}: \doi{10.1038/s41598-025-33057-9}}
}

@article{abdullah2026rhythmx,
  title={RhythmX™: An Interpretable Self-Supervised Contrastive Learning Framework for Heartbeat Classification},
  author={Abdullah and Fatima, Zulaikha and Safder, Haris Ali and Manzoor, Mubasher and S{\'a}nchez-Mejorada, Carlos Guzm{\'a}n and Torres Ruiz, Miguel Jes{\'u}s and Quintero T{\'e}llez, Rolando},
  journal={Technologies},
  volume={14},
  number={3},
  pages={148},
  year={2026},
  publisher={MDPI},
  note    = {{DOI}: \doi{10.3390/technologies14030148}}
}

@article{tiffany2026robust,
  title={Robust multilabel ECG classification under class imbalance using augmentation and adaptive iterative training},
  author={Tiffany, Joshua and Zhang, Qi},
  journal={Biomedical Signal Processing and Control},
  volume={118},
  pages={109767},
  year={2026},
  publisher={Elsevier},
  note    = {{DOI}: \doi{10.1016/j.bspc.2026.109767}}
}

@article{ma2026self,
  title={Self-supervised contrastive learning achieves 12-lead ECG classification},
  author={Ma, Ke and Zhang, Tao and Zhang, Hengyuan and Huang, Wu},
  journal={Biomedical Signal Processing and Control},
  volume={112},
  pages={108420},
  year={2026},
  publisher={Elsevier},
  note    = {{DOI}: \doi{10.1016/j.bspc.2025.108420}}
}

@article{yuan2023target,
  title={Target-oriented augmentation privacy-protection domain adaptation for imbalanced ECG beat classification},
  author={Yuan, Liqiang and Siyal, Mohammed Yakoob},
  journal={Biomedical Signal Processing and Control},
  volume={86},
  pages={105308},
  year={2023},
  publisher={Elsevier},
  note    = {{DOI}: \doi{10.1016/j.bspc.2023.105308}}
}

@article{chen2025temporal,
  title={Temporal and spatial self supervised learning methods for electrocardiograms},
  author={Chen, Wenping and Wang, Huibin and Zhang, Lili and Zhang, Min},
  journal={Scientific Reports},
  volume={15},
  number={1},
  pages={6029},
  year={2025},
  publisher={Nature Publishing Group UK London},
  note    = {{DOI}: \doi{10.1038/s41598-025-90084-2}}
}

@article{chatterjee2026toward,
  title={Toward robust automated cardiovascular arrhythmia detection using self-supervised learning and 1-dimensional vision transformers},
  author={Chatterjee, Mitchell and Chan, Adrian DC and Komeili, Majid},
  journal={Scientific Reports},
  volume={16},
  number={1},
  pages={11793},
  year={2026},
  publisher={Nature Publishing Group UK London},
  note    = {{DOI}: \doi{10.1038/s41598-026-41549-5}}
}

@article{weimann2021transfer,
  title={Transfer learning for ECG classification},
  author={Weimann, Kuba and Conrad, Tim OF},
  journal={Scientific reports},
  volume={11},
  number={1},
  pages={5251},
  year={2021},
  publisher={Nature Publishing Group UK London},
  note    = {{DOI}: \doi{10.1038/s41598-021-84374-8}}
}

@article{ayyub2025comprehensive,
  title={A comprehensive review of AI-based detection of arrhythmia using electrocardiogram (ECG)},
  author={Ayyub, Ahtisham and Politis, Christos and Usman, Muhammad Arslan},
  journal={Computers in Biology and Medicine},
  volume={196},
  pages={110594},
  year={2025},
  publisher={Elsevier},
  note    = {{DOI}: \doi{10.1016/j.compbiomed.2025.110594}}
}

@article{soltanieh2023distribution,
  title={In-distribution and out-of-distribution self-supervised ecg representation learning for arrhythmia detection},
  author={Soltanieh, Sahar and Hashemi, Javad and Etemad, Ali},
  journal={IEEE Journal of Biomedical and Health Informatics},
  volume={28},
  number={2},
  pages={789--800},
  year={2023},
  publisher={IEEE},
  note    = {{DOI}: \doi{10.1109/JBHI.2023.3331626}}
}

@article{imtiaz2024cross,
  title={Cross-database and cross-channel electrocardiogram arrhythmia heartbeat classification based on unsupervised domain adaptation},
  author={Imtiaz, Md Niaz and Khan, Naimul},
  journal={Expert Systems with Applications},
  volume={244},
  pages={122960},
  year={2024},
  publisher={Elsevier},
  note    = {{DOI}: \doi{10.1016/j.eswa.2023.122960}}
}

@inproceedings{ballas2022domain,
  title={A domain generalization approach for out-of-distribution 12-lead ecg classification with convolutional neural networks},
  author={Ballas, Aristotelis and Diou, Christos},
  booktitle={2022 IEEE Eighth International Conference on Big Data Computing Service and Applications (BigDataService)},
  pages={9--13},
  year={2022},
  organization={IEEE},
  note    = {{DOI}: \doi{10.1109/BigDataService55688.2022.00009}}
}

@article{mehari2022self,
  title={Self-supervised representation learning from 12-lead ECG data},
  author={Mehari, Temesgen and Strodthoff, Nils},
  journal={Computers in biology and medicine},
  volume={141},
  pages={105114},
  year={2022},
  publisher={Elsevier},
  note    = {{DOI}: \doi{10.1016/j.compbiomed.2021.105114}}
}

@article{hong2020opportunities,
  title={Opportunities and challenges of deep learning methods for electrocardiogram data: A systematic review},
  author={Hong, Shenda and Zhou, Yuxi and Shang, Junyuan and Xiao, Cao and Sun, Jimeng},
  journal={Computers in biology and medicine},
  volume={122},
  pages={103801},
  year={2020},
  publisher={Elsevier},
  note    = {{DOI}: \doi{10.1016/j.compbiomed.2020.103801}}
}

@article{silva2026systematic,
  title={A systematic review of ECG arrhythmia classification: embedded feasibility, adherence to standards, and fair evaluation},
  author={Silva, Guilherme AL and Silva, Pedro HL and Moreira, Gladston JP and Freitas, Vander L de S and Gertrudes, Jadson C and Luz, Eduardo J da S},
  journal={Research on Biomedical Engineering},
  volume={42},
  number={3},
  pages={42},
  year={2026},
  publisher={Springer},
  note    = {{DOI}: \doi{10.1007/s42600-026-00485-z}}
}

@inproceedings{kachuee2018ecg,
  title={Ecg heartbeat classification: A deep transferable representation},
  author={Kachuee, Mohammad and Fazeli, Shayan and Sarrafzadeh, Majid},
  booktitle={2018 IEEE international conference on healthcare informatics (ICHI)},
  pages={443--444},
  year={2018},
  organization={IEEE},
  note    = {{DOI}: \doi{10.1109/ICHI.2018.00092}}
}

@inproceedings{cao2023ecg,
  title={ECG Heartbeat classification using deep transfer learning with Convolutional Neural Network and STFT technique},
  author={Cao, Minh and Zhao, Tianqi and Li, Yanxun and Zhang, Wenhao and Benharash, Peyman and Ramezani, Ramin},
  booktitle={Journal of Physics: Conference Series},
  volume={2547},
  number={1},
  pages={012031},
  year={2023},
  organization={IOP Publishing},
  note    = {{DOI}: \doi{10.1088/1742-6596/2547/1/012031}}
}

@article{mckeen2025ecg,
  title={Ecg-fm: An open electrocardiogram foundation model},
  author={McKeen, Kaden and Masood, Sameer and Toma, Augustin and Rubin, Barry and Wang, Bo},
  journal={Jamia Open},
  volume={8},
  number={5},
  pages={ooaf122},
  year={2025},
  publisher={Oxford University Press},
  note    = {{DOI}: \doi{10.1093/jamiaopen/ooaf122}}
}

@article{de2004automatic,
  title={Automatic classification of heartbeats using ECG morphology and heartbeat interval features},
  author={De Chazal, Philip and O'Dwyer, Maria and Reilly, Richard B},
  journal={IEEE transactions on biomedical engineering},
  volume={51},
  number={7},
  pages={1196--1206},
  year={2004},
  publisher={IEEE},
  note    = {{DOI}: \doi{10.1109/TBME.2004.827359}}
}

@article{vaswani2017attention,
  title={Attention is all you need},
  author={Vaswani, Ashish and Shazeer, Noam and Parmar, Niki and Uszkoreit, Jakob and Jones, Llion and Gomez, Aidan N and Kaiser, {\L}ukasz and Polosukhin, Illia},
  journal={Advances in neural information processing systems},
  volume={30},
  year={2017},
  note    = {{DOI}: \doi{10.48550/arXiv.1706.03762}}
}

@inproceedings{chen2020simple,
  title={A simple framework for contrastive learning of visual representations},
  author={Chen, Ting and Kornblith, Simon and Norouzi, Mohammad and Hinton, Geoffrey},
  booktitle={International conference on machine learning},
  pages={1597--1607},
  year={2020},
  organization={PmLR},
  note    = {{DOI}: \doi{10.48550/arXiv.2002.05709}}
}

@misc{incart,
    author = "Evgeny Yakushenko" ,
    title = "St Petersburg INCART 12-lead Arrhythmia Database" ,
    year = 2008,
    url = "https://physionet.org/content/incartdb/1.0.0/" , 
    addendum = "(accessed: 01.09.2023)",
  note    = {{DOI}: \doi{}}
}

@article{moody2001impact,
  title={The impact of the MIT-BIH arrhythmia database},
  author={Moody, George B and Mark, Roger G},
  journal={IEEE engineering in medicine and biology magazine},
  volume={20},
  number={3},
  pages={45--50},
  year={2001},
  publisher={IEEE},
  note    = {{DOI}: \doi{10.1109/51.932724}}
}

@book{greenwald1990improved,
  title={Improved detection and classification of arrhythmias in noise-corrupted electrocardiograms using contextual information},
  author={Greenwald, Scott David and Patil, Ramesh S and Mark, Roger G},
  year={1990},
  publisher={IEEE},
  note    = {{DOI}: \doi{10.1109/CIC.1990.144257}}
}

@misc{AAMI_EC57_2012,
  author       = {{Association for the Advancement of Medical Instrumentation}},
  title        = {{ANSI/AAMI EC57:2012/(R)2020: Testing and reporting performance results of cardiac rhythm and ST segment measurement algorithms}},
  year         = {2012},
  address      = {Arlington, VA, USA},
  note    = {{DOI}: \doi{10.2345/9781570204784.ch1}}
}

@article{chicco2020advantages,
  title={The advantages of the Matthews correlation coefficient (MCC) over F1 score and accuracy in binary classification evaluation},
  author={Chicco, Davide and Jurman, Giuseppe},
  journal={BMC genomics},
  volume={21},
  number={1},
  pages={6},
  year={2020},
  publisher={Springer},
  note    = {{DOI}: \doi{10.1186/s12864-019-6413-7}}
}

@article{kvaalseth1989note,
  title={Note on Cohen's kappa},
  author={Kv{\aa}lseth, Tarald O},
  journal={Psychological reports},
  volume={65},
  number={1},
  pages={223--226},
  year={1989},
  publisher={SAGE Publications Sage CA: Los Angeles, CA},
  note    = {{DOI}: \doi{10.2466/pr0.1989.65.1.223}}
}

\end{document}